%% file: main.tex
\documentclass[sigconf]{acmart}

\usepackage{lscape}
\usepackage{multirow}
\usepackage{subcaption}

\usepackage{xcolor}
\definecolor{royalblue}{rgb}{0.25, 0.41, 0.88}
\newcommand*{\editing}{\textcolor{black}}

\AtBeginDocument{%
  }

\copyrightyear{2026}
\acmYear{2026}
\setcopyright{cc}
\setcctype{by}
\acmConference[FAccT '26]{Proceedings of the 2026 ACM Conference on Fairness, Accountability, and Transparency}{June 25--28, 2026}{Montr\'eal, Canada}
\acmBooktitle{Proceedings of the 2026 ACM Conference on Fairness, Accountability, and Transparency (FAccT '26), June 25--28, 2026, Montr\'eal, Canada}
\acmPrice{}
\acmDOI{XXXXXXX.XXXXXXX}
\acmISBN{}

\begin{document}

\title{Bureaucratic Silences: What the Canadian AI Register Reveals, Omits, and Obscures}

\author{Dipto Das}
\affiliation{%
  \department{Department of Computer Science}
  \institution{University of Toronto}
  \city{Toronto}
  \state{Ontario}
  \country{Canada}
}
\email{dipto.das@utoronto.ca}

\author{Christelle Tessono}
\affiliation{%
  \department{Faculty of Information}
  \institution{University of Toronto}
  \city{Toronto}
  \state{Ontario}
  \country{Canada}
}
\email{christelle.tessono@mail.utoronto.ca}

\author{Syed Ishtiaque Ahmed}
\affiliation{%
  \department{Department of Computer Science}
  \institution{University of Toronto}
  \city{Toronto}
  \state{Ontario}
  \country{Canada}
}
\email{ishtiaque@cs.toronto.edu}

\author{Shion Guha}
\affiliation{%
  \department{Faculty of Information}
  \institution{University of Toronto}
  \city{Toronto}
  \state{Ontario}
  \country{Canada}
}
\email{shion.guha@utoronto.ca}



\begin{abstract}
  In November 2025, the Government of Canada operationalized its commitment to transparency by releasing its first Federal AI Register. In this paper, we argue that such registers are not neutral mirrors of government activity, but active instruments of \textbf{ontological design} that configure the boundaries of accountability. We analyzed the Register's complete dataset of 409 systems using the Algorithmic Decision-Making Adapted for the Public Sector (ADMAPS) framework, combining quantitative mapping with deductive qualitative coding. Our findings reveal a sharp divergence between the rhetoric of ``sovereign AI" and the reality of bureaucratic practice: while 86\% of systems are deployed internally for efficiency, the Register systematically obscures the human discretion, training, and uncertainty management required to operate them. By privileging technical descriptions over sociotechnical context, the Register constructs an ontology of AI as ``reliable tooling" rather than ``contestable decision-making." We conclude that without a shift in design, such transparency artifacts risk automating accountability into a performative compliance exercise, offering visibility without contestability.
\end{abstract}

\maketitle
\input{sections/introduction}
\input{sections/literature_review}
\input{sections/methods}
\input{sections/results}
\input{sections/discussion}
\input{sections/conclusion}

\bibliographystyle{ACM-Reference-Format}
\bibliography{main}

\end{document}

%% file: sections/introduction.tex
\section{Introduction}
In November 2025, the Government of Canada (GC) publicly released its first government-wide artificial intelligence (AI) Register, disclosing over 400 AI systems used across more than 40 federal institutions~\cite{steven2025ottawa}. The GC highlighted making ``public service more efficient" as the motivation for incorporating AI into federal government operations, for which this register would reduce duplication, and help departments identify opportunities to work more efficiently~\cite{steven2025ottawa}. Various geopolitical entities are building similar registers~\cite{algorithmwatch2025ai, murad2021beyond, haataja2020public, kaushal2024automated}, such as the European Union (EU)'s Digital Services Act (DSA), positioning such disclosure infrastructures as a central mechanism for governing algorithmic systems used in the public sector. From the algorithmic fairness, accountability, and transparency (FAccT) research perspective, we ask: \emph{how meaningfully do AI registers enable accountability, or do they merely institutionalize its appearance?}

FAccT research has largely framed disclosure, through documentation standards, audits, and reporting requirements, as a necessary condition for oversight~\cite{gpai2024algorithmic}. While these approaches have advanced important norms (e.g., datasheet, model cards)~\cite{gebru2021datasheets, mitchell2019model}, critical scholarship~\cite{selbst2019fairness} has shown that institutional factors (e.g., incentives, structures) substantially shape disclosure efforts, such as AI registers. Moreover, emerging discourse on AI governance often treats AI systems as conceptually interchangeable, increasingly conflating generative and large language model (LLM)–based systems~\cite{taeihagh2025governance}. This emphasis on highly visible and popular tools obscures the continued operation of long-standing predictive, scoring, and decision-support systems that continue to shape public-sector decision-making~\cite{zuiderwijk2021implications}. By rendering these systems visible and classifying them in particular ways, AI registers shape how accountability is practiced, and more significantly, define what counts as AI in governance practice. 

Rather than asking what information the Canadian AI Register discloses, we examine how it delineates responsibility, represents uncertainty, and locates discretion. To do so, we used the algorithmic decision-making adapted for the public sector (ADMAPS) framework~\cite{saxena2021framework}, which foregrounds various types of interactions within public-sector governance. Thus, this paper makes three contributions. First, we provide an empirical analysis of the complete Canadian federal Public AI Register to systematically examine 409 documented systems. Second, we introduce the concept of \emph{bureaucratic silences} to describe how AI registers structure what is disclosed about public sector AI systems (e.g., tool- and developer-level disclosures, technical capabilities, and efficiency), and what remains illegible or off the record (e.g., human, bureaucratic, and contextual factors), particularly around discretion, infrastructure, and uncertainty. As similar registers emerge across Europe and beyond~\cite{kaushal2024automated, haataja2020public, murad2021beyond}, the silences we identify and problematize should be examined as systemic properties of register-driven transparency regimes, rather than uniquely Canada-specific peculiarities--an insight that broadens the relevance of our analysis for global AI governance. Third, we argue that AI registers should be understood as instruments of \emph{ontological design}, which shape how accountability is defined and enacted, with concrete implications for how future registers should be designed if they are to support meaningful democratic oversight and public trust in public-sector AI. We situate our findings by discussing how the AI register contributes to and articulates Canada's preparedness for its national AI governance and sovereignty strategies.


%% file: sections/literature_review.tex
\section{Literature Review}\label{sec:literature_review}
Our literature review brings together three bodies of work that frame how algorithmic accountability is enacted in practice: sociotechnical perspectives on accountability and transparency, research on documentation and disclosure as governance infrastructures, and critical studies of public-sector algorithmic systems. We draw on these strands to motivate our choice of ADMAPS as the analytical lens for examining the \editing{Government of Canada (GC)} AI Register.

\subsection{Algorithmic Accountability and Transparency in Sociotechnical Systems}
Accountability and transparency have emerged as central concepts in scholarship on the social impacts of algorithmic and AI systems. Whereas social sciences understand accountability as the everyday person's ability to hold an institution responsible~\cite{lindberg2013mapping}, from a technical standpoint, accountability refers to the capacity of algorithmic systems to trace, justify, and assign responsibility for their decisions and outcomes~\cite{binns2018algorithmic, diakopoulos2016accountability, wieringa2020account}, whereas transparency is commonly understood as the availability of information that enables explanation, documentation, or disclosure of system properties~\cite{balasubramaniam2023transparency, larsson2020transparency, valderrama2023state}. These concepts are often operationalized through technical and procedural interventions, such as explainable models, documentation frameworks, audit mechanisms, and reporting standards, which aim to render algorithmic systems more legible to developers, regulators, and affected users~\cite{mitchell2019model, sokol2020explainability, felzmann2019transparency}.

Beyond such technical understandings of algorithmic accountability and transparency, scholars increasingly conceptualize them as shaped by sociocultural contexts, institutional logics, organizational structures, legal mandates, political priorities, and power asymmetries~\cite{selbst2019fairness, ananny2018seeing, suchman2002located, das2021jol}. Within this line of critique, accountability is reframed as a question of governance: who is accountable to whom, through what mechanisms, and under what institutional conditions~\cite{binns2018algorithmic, wieringa2020account}. For example, studies have shown that algorithmic supply chains fragment responsibility across vendors, developers, and deploying institutions, complicating traditional notions of oversight and liability~\cite{cobbe2023understanding}. In such a scenario, where responsibility for algorithmic decisions is distributed across organizational roles, infrastructures, and decision points rather than residing solely in models or code, traceability emerges as a principle for operationalizing accountability~\cite{kroll2021outlining}. In this governance-oriented view, transparency is not reducible to the disclosure of technical details but is better understood as a ``communicative constellation" and a form of ``disclosure by design," in which disclosures are situated, audience-specific, and shaped by institutional incentives, actively structuring how information is interpreted, which forms of accountability become possible, and who is empowered to act~\cite{eyert2023rethinking,norval2022disclosure}.

Critical scholars have further highlighted the organizational stakes of accountability and transparency practices themselves, such as how disclosure datasets embody politics in defining what counts as accountable knowledge~\cite{poirier2022accountable, boag2022tech}. They showed how accountability is often pursued through collective action \editing{(e.g., advocacy)} as a negotiation between efforts to strengthen accountability through disclosure and attempts to reduce regulatory burden. Algorithmic accountability depends on alignment among system design, professional judgment, and organizational context~\cite{veale2018fairness}, and is an ongoing practice rather than a one-time technical intervention~\cite{johnson2021algorithmic}. \editing{Redden and colleagues have examined its extent in the public sector through case studies in fraud detection, child welfare, social services, and policing~\cite{dencik2018data, redden2020datafied, redden2022datafied} across Europe, North America, and Australia~\cite{redden2022automating}. In the Canadian context, Redden~\cite{redden2018democratic} similarly investigated government discourses and practices surrounding big data adoption in the public sector using counter-mapping methods and freedom-of-information requests. These studies demonstrated significant gaps in existing transparency regimes since algorithmic governance is often difficult to trace due to fragmented documentation practices, institutional opacity, and limited proactive disclosure.} Overall, accountability is a relational and institutional practice that requires examining the documentation norms, disclosure regimes, and governance artifacts. \editing{Hence, researchers have called for systematic documentation and public disclosure mechanisms that enable oversight of algorithmic systems in governance.}

\subsection{Accountability in Governance through Documentation and Disclosure}
\editing{Sociotechnically grounded AI governance research} highlights how documentation and disclosure, operating at the intersection of technical practice and organizational governance, translate abstract accountability commitments into standardized, inspectable artifacts that circulate across organizations, regulators, and publics\editing{~\cite{gebru2021datasheets, winecoff2025improving, ojewale2026audit}}. Examples of these artifacts are documentation frameworks such as datasheets and model cards, which promote responsible AI development by standardizing disclosures about provenance, intended use, limitations, and risks. These frameworks also function as community practices, embedding disciplinary norms, institutional incentives, and implicit assumptions about responsibility and audience into their design and use~\cite{mcmillan2024data, dergacheva2023one}. However, documentation does not simply describe underlying systems--instead, it actively shapes which forms of information are legible and what kinds of accountability claims become possible. Therefore, data systems' accompanying documentation must be ``read" critically: what decisions they encode about classification and how those would shape downstream interpretation and governance~\cite{poirier2021reading, kuehnert2025and}.

Documentation practices not only support transparency and accountability but also delimit their scope by foregrounding certain risks, actors, and values while understating others. Claims about scale, representativeness, or neutrality often become central to these documentations and obscure the contingent social and organizational conditions under which data are produced and maintained~\cite{kitchin2016makes}. For instance, scholars have traced how the discourse around data systems stabilizes narratives~\cite{denton2021genealogy} by shaping how influential benchmarks are conceptualized, reducing concerns about fairness and accountability to concerns about sufficient data~\cite{aragon2022human, jarrahi2023principles}, and by making certain types of data labor invisible~\cite{scheuerman2025data, sen2015turkers}.

Beyond individual datasets or models, normative demands for accountability and transparency are often enacted through formalized governance infrastructures and structured reporting obligations. \editing{AI registers have emerged as a governance instrument in several jurisdictions seeking to increase transparency around public-sector AI systems. Municipal governments such as Amsterdam~\cite{murad2021beyond} and Helsinki~\cite{haataja2020public} were among the earliest adopters of public algorithm registers, publishing structured documentation about the AI systems used in public services, which have since inspired similar calls for disclosure mechanisms in other jurisdictions, such as the EU's DSA~\cite{klafkowska5257655search} Bangladesh's AI policy~\cite{govbd2024national}, and Canada's municipal and organizational AI registers~\cite{sieber2026building, sengupta2022algorithm}. Empirical studies focusing on these transparency databases, AI registers, ethical charters, and compliance repositories across various levels, scales, and contexts have} examined how documenting patterns in self-reported actions, inconsistencies in categorization, and limits to contextual explanation~\cite{trujillo2025dsa, drolsbach2024content, kelly2022facial} standardize disclosure pipelines and moderation practices that prioritize structured, machine-readable reporting over interpretive openness~\cite{kaushal2024automated}. Scoping these insights in the context of the public sector, researchers have analyzed how disclosure templates, reporting thresholds, and organizational incentives shape what institutions can or choose to document about deployed systems~\cite{nieuwenhuizen2024algorithm, hogberg2024stabilizing}, \editing{similar to Redden et al.'s insights~\cite{redden2018democratic, redden2022automating} we discussed earlier}. Because of these design choices, accountability initiatives often fall short of their regulatory objectives~\cite{groesch2025big}. These challenges often exacerbate in Global South contexts \editing{due to different} AI ethics that underpin these policies~\cite{bengio2025international, okolo2023ai, mayeesha2024ai4bangladesh}.

Related works on AI ethics charters and manifestos map how normative voluntary disclosures and commitments across institutions produce an appearance of ethical accountability at a meso-governance level, without necessarily mandating enforceable accountability~\cite{gornet2024mapping, hagendorff2020ethics, correa2023worldwide}. Taken together, this body of work traces a shift from transparency as an aspirational principle toward transparency as an infrastructural practice. Similarly, studies of large-scale data infrastructures, like Common Crawl, demonstrate how disclosure practices can obscure responsibility by diffusing accountability across actors, licenses, and downstream users~\cite{baack2024critical}. In the context of fairness and accountability research, such ``lazy" data documentation practices can entrench methodological shortcuts and epistemic blind spots, reinforcing the appearance of accountability without substantively addressing underlying harms~\cite{simson2024lazy}.

\subsection{Critical Analysis of Accountability of Algorithmic Systems in the Public Sector}
Algorithmic accountability studies employ a wide set of empirical approaches. A prominent method among these is auditing, which examines decision-making patterns while varying inputs to surface bias, error, or harm~\cite{sandvig2014auditing, raji2020closing}. Both researchers and everyday users have used algorithmic audits to scrutinize AI systems~\cite{metaxa2021auditing, shen2021everyday, deng2023understanding}, like recommendation and search algorithms~\cite{baeza2020bias, robertson2018auditing}, computer vision-based processes (e.g., generative art~\cite{srinivasan2021biases}, image captioning~\cite{zhao2021understanding}, facial recognition~\cite{buolamwini2018gender}), and language technologies (e.g., sentiment analysis~\cite{kiritchenko-mohammad-2018-examining, das2024colonial}, hate-speech detector~\cite{sap2019risk}, machine translation~\cite{savoldi-etal-2022-morphosyntactic}, text generation~\cite{fan-gardent-2022-generating}). Scholars have also emphasized that deciphering the algorithmic outputs alone is often insufficient for rigorous accountability analysis~\cite{gansky2022counterfacctual}, particularly when system behavior is shaped by organizational workflows, data pipelines, and governance constraints that remain undocumented or inaccessible~\cite{casper2024black, mcconvey2024not, saxena2024algorithmic}. In such cases, investigative and journalistic approaches, combining technical inspection with document analysis, interviews, and policy tracing, are also effective in identifying opaque practices and institutional accountability gaps~\cite{diakopoulos2015algorithmic, diakopoulos2025prospective}.

While impact assessments structure accountability by anticipating and documenting potential harms, trade-offs, and mitigation strategies, their effectiveness depends on institutional incentives and implementation practices~\cite{watkins2021governing}. In public services, accountability challenges are amplified by high stakes, legal mandates, and complex procurement arrangements~\cite{brown2019toward}. To navigate through these, public-sector algorithms often redistribute discretion rather than eliminate it, shifting responsibility across frontline workers, technical systems, and administrative hierarchies~\cite{veale2018fairness, saxena2023rethinking, moon2025datafication}. \editing{In the Canadian context, the Directive on Automated Decision-Making~\cite{gc2021adm}, introduced by the Treasury Board Secretariat in 2019, shapes such algorithmic governance practices. The directive requires federal agencies deploying automated decision systems to complete an Algorithmic Impact Assessment (AIA)~\cite{gc2026aia} that evaluates potential impacts of algorithmic decision-making on rights, fairness, and administrative accountability. While the AIA functions primarily as an ex-ante assessment mechanism, the GC AI Register serves as a disclosure infrastructure, which documents and describes systems in development or already deployed across federal agencies.}

To critically examine the GC AI Register, we draw on the algorithmic decision-making adapted for the public sector (ADMAPS) framework~\cite{saxena2021framework}. \editing{Because it focuses on how algorithmic systems are embedded within bureaucratic processes, institutional practices, and data infrastructures rather than treating them solely as technical artifacts, the ADMAPS framework offers a useful analytical lens for our study.} Unlike actor/stage mappings that catalog governance instruments~\cite{kuehnert2025and} or tools for AI impact assessment in terms of prospective risks~\cite{johnson2023assessing}, ADMAPS offers an analytical lens for reading the GC AI Register as an accountability artifact embedded in the public sector. This framework conceptualizes algorithmic governance as a sociotechnical arrangement shaped by the interaction of human discretion, bureaucratic processes, and algorithmic decision-making. Human discretion encompasses professional expertise, value judgments, and heuristic decision-making exercised by frontline practitioners. Bureaucratic processes capture organizational conditions such as resources and constraints, administrative routines and training, and governing laws and policies. Algorithmic decision-making attends to the role of data, the type of decision support provided (e.g., predictive or prescriptive), and the degree of uncertainty embedded in algorithmic systems. Overall, this offers a structured lens for analyzing how the GC AI Register defines, represents, organizes, and assesses the AI systems across various departments and agencies of the Canadian government.


%% file: sections/methods.tex
\section{Methods}
In this section, we explain our mixed-methods approach to analyzing how AI systems are framed in the GC AI Register, combining mixed-method exploration with ADMAPS-guided critical discourse analysis.

\subsection{Data Collection, Preprocessing, and Exploratory Analysis}
Our study draws on the GC AI Register~\cite{gc2025register}, which provides structured, inventory-like information on 409 AI systems, ranging from early initiatives to fully operational tools supporting service delivery and internal operations across 42 federal departments and agencies. Each entry includes fields, such as: ID, AI system's name, description, primary users, status, date, capabilities, data sources, results, as well as which government organization is associated with it, who developed it, and vendor information, and whether and which personal information banks are involved with it, in both English and French. Our analysis focuses on how public-sector AI systems are represented and documented in the register, rather than on auditing the registered AI systems or offering an ethnographic account of their day-to-day use. In other words, our object of study is the register as a sociotechnical governance artifact that projects particular views of data, decision-making, and accountability, in line with the ADMAPS dimensions. In this paper, we treat the register not only as an administrative dataset but also as an institutional governance artifact that reflects how AI use is documented, framed, and communicated in the context of public-sector policy commitments and accountability norms.

In the tabular dataset, each row corresponds to a unique AI system. We retained only the English columns and added a column for the acronyms of the corresponding government organizations, as listed in~\cite{gc2025departments}. We formatted the categorical fields for consistency (e.g., capitalization and spacing) and imputed empty categorical fields as `Unspecified'. We exploratorily analyzed the quantitative distributions of nominal variables: primary users, associated government organization, development status, vendor information, and the use of personal information banks. To examine patterns in reported AI capabilities, we applied lightweight text analysis to the capability descriptions after removing stop words and common domain-generic terms (e.g., `model,' `data,' `AI'), extracting tokens, bigrams, and trigrams, and generating a word cloud to surface recurrent capability themes across systems in the register.
\subsection{Data Analysis}
We used deductive qualitative analysis guided by the ADMAPS framework~\cite{saxena2021framework}. Using it as the analytical lens, the first author conducted the initial open coding, i.e., identified recurring representational patterns: entities, tasks, objectives, etc., that frequently appeared in the textual descriptions of AI systems and their results within predefined theoretical dimensions. Using each AI system entry in the Register as the unit of analysis, the codebook was iteratively developed: the codes generated by the first author were refined through discussions with the \editing{other three authors}, where disagreements or ambiguities were \editing{documented in analytic memos and} collaboratively resolved, and codes were merged, split, or adjusted to maintain consistency. \editing{Following social computing research guidelines for such approaches~\cite{mcdonald2019reliability}, we did not calculate the inter-coder reliability score for the codes. To examine} how AI systems are framed, \editing{we mapped the codes} in relation to the ADMAPS dimensions: algorithmic decision-making, bureaucratic procedure, and human discretion~\cite{saxena2021framework}, explained earlier in Section~\ref{sec:literature_review}. Drawing on critical discourse analysis that considers what is said, emphasized, and left unsaid in texts \editing{and documents~\cite{meyer2001methods, fairclough2023critical}, we examine the register as an institutional governance artifact whose standardized descriptions and categories structure how AI systems and accountability relations are represented. Accordingly, we treated both recurring emphases and systematic} silences as analytically meaningful in our iterative coding. For example, while coding for degrees of uncertainty, we noted explicit references to pilots or experimentation, as well as the absence of such references in descriptions of mature or operational systems. We juxtaposed exploratory quantitative insights with qualitative interpretation (e.g., considering frequency as an indicator of institutional normalization) to understand how public-sector AI systems are framed within federal governance.

\subsection{Limitations}
This study has several limitations. First, our analysis is bounded by what is documented in the register and cannot access undocumented systems, informal workarounds, or the situated practices of frontline workers. Hence, our study illuminates the politics of disclosure and representation rather than the operational performance or impacts of each system. Second, our coding approach is interpretive and theory-driven, anchored in ADMAPS. While other frameworks could foreground different aspects of the same entries, our objective is to critically examine the Canadian federal AI systems with reference to their bureaucratic situatedness and institutional logics.


%% file: sections/results.tex
\section{Results}
We organize our results through the ADMAPS framework, examining (1) how algorithmic decision-making is represented and operationalized; (2) how AI systems are described in relation to bureaucratic processes; and (3) how human discretion, expertise, and heuristics are reconfigured. Across each category, we examined its corresponding elements.

\subsection{Algorithmic Decision-Making}
In this section, we analyze how algorithmic decision-making is represented in the GC AI register by identifying the type of data, the forms of decision-making support, and the degrees of uncertainty embedded in the algorithmic systems. 

\subsubsection{Relevant Data}\label{sec:relevant_data}
The AI systems cataloged in the GC AI Register draw on an uneven landscape of data sources: less than a quarter use personal information (21.4\%), and 24.4\% do not specify their data practices. We identified several data types and sources in the AI register: (a) Administrative and transactional data, (b) Open and public-facing web data, (c) Commercial/proprietary data, and (d) User-generated input.

Many AI systems rely heavily on \textbf{administrative records and transactional data}, including case files, enforcement logs, inspection reports, adjudication notes, ticketing histories, and sensor-mediated event logs. These sources are often deeply tied to bureaucratic routines and frontline documentation practices. Examples include the Canada Border Services Agency (CBSA)'s Fuzzy Search/SSAName3, a tool that flags high-risk travelers and goods. It operates across \emph{``multiple different datasets so they are properly indexed, identified, scored, and ranked"}. In fact, SSAName3 uses 28 different personal data banks, the most of any registered AI system. Similarly, the Canada Revenue Agency's Anomaly Detection AI system for detecting fraudulent transactions uses billions of historical records and user access logs. Across these contexts, administrative data act as frozen bureaucratic interpretations, not objective realities. They risk anticipating prior patterns to recur or exacerbating what the system already fails to see, especially when models become authoritative over the very workers who originally created these records. Moreover, many system descriptions specify only \emph{``internal documents," or ``data from SharePoint,"} without clarifying provenance, governance, or structure.

The second category comprises {\textbf{open and public-facing web data}}. These sources are curated for external communication with the public rather than for internal policy execution. For example, AgPal Chat relies on the AgPal website to surface \emph{``information about more than 400 federal, provincial, territorial, and municipal programs and services."} Similarly, the Canadian Trade Commissioner Service (TCS) website chatbot, referred to as the Digital Experience Platform, uses the TCS website as its primary knowledge base, including \emph{``50 bilingual frequently asked questions."}

The third category draws on \textbf{commercial datasets and proprietary corpora}, typically developed and maintained by vendors. For instance, Westlaw-powered tools in the GC AI register use models trained on \emph{``relationships within its legal corpus,"} and \emph{``all content is maintained within the Westlaw platform."} Similar patterns appear in vendor-built systems such as Darktrace, used by for the Canadian Grain Commission and the Automated Prohibited Item Detection (APIDS) system for the Canadian Air Transport Authority, where data sources and users are minimally specified. While some degree of vagueness may be intentional, given the system's security-critical role, it raises questions about whether its users have operational visibility into the data. 


The fourth category of data draws on \textbf{user-generated and submitted content} such as uploaded documents, conversational queries, scanned IDs, interview responses, and long-form narrative explanations. Systems such as CBSA's Client Reporting and Engagement System, Employment managed by Employment and Social Development Canada (ESDC)'s Record of Employment, and AI-assisted medical and immigration assessments used by Immigration, Refugees, and Citizenship Services Canada (IRCC), rely on highly heterogeneous, context-rich inputs that encode emotional, linguistic, and cultural variation. When optimized for structured analysis, AI systems may flatten ambiguity, overfit superficial patterns, and reduce interpretive flexibility. Although efficiency gains are emphasized, the register offers little insight into how human adjudicators intervene to compensate for these limits. 

\subsubsection{Types of Decision-Support}
\begin{figure}[!ht]
    \centering
    \includegraphics[width=\linewidth]{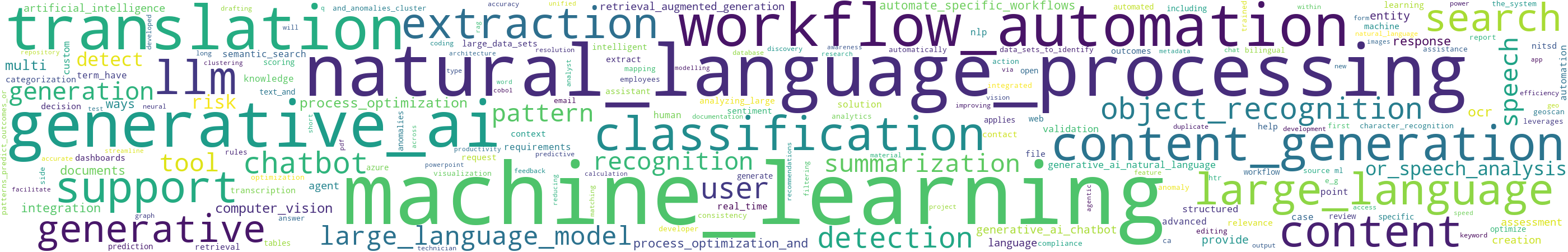}
    \caption{A word cloud of the technical capabilities of the AI systems.}
    \label{fig:capabilities}
\end{figure}

The GC AI register identified the technical underpinnings of many AI systems, such as natural language processing (NLP), LLM, machine learning (ML), computer vision, and optical character recognition (OCR). Figure~\ref{fig:capabilities} shows that NLP, LLM, and Generative AI are the most commonly reported technical capabilities. Using the ADMAPS framework, we identify four categories of decision-support across the register:  (a) informational, (b) predictive, (c) prescriptive, and (d) operational automation, spanning functions such as summarization, estimation, recommendation, and optimization.

A large share of the AI systems on the register provide \textbf{informational support}, such as retrieval, summarization, document generation, or conversational access to knowledge repositories. While these systems do not make decisions, they shape how insights are shared with government organizations, thereby indirectly influencing discretion and workflow. For example, the Canadian Radio-Television and Telecommunications Commission (CRTC)'s CANChat built on LLMs, generative AI and chatbot functionality--has been adopted or adapted by several other departments (e.g., Indigenous Services, Services Canada, Library and Archives Canada, Shared Services Canada), while similar NLP and LLM based tools enable users to \emph{``chat with"} documents or generate transcripts and summaries from audio. By shifting discretion from interpretation to verification, these systems subtly standardize interpretive practices and narrow the range of perspectives in casework and policy analysis.  


We found that several AI systems in the register satisfy the criteria for which the ADMAPS framework calls \textbf{prescriptive systems}, which shape the process or structure of bureaucratic response by recommending or directing actions as well as by triaging and routing individuals and cases. For example, ESDC uses an AI system that highlights \emph{``essential details"} for Labour Market Impact Assessment\footnote{\editing{This} is a document that an employer in Canada may need to get before hiring a foreign worker~\cite{gc2025lmia}. A positive \editing{assessment implies} that there is a need for a foreign worker to fill the job.} to guide officers reviewing Temporary Foreign Worker Program applications. Likewise, CBSA's Security Screening Automation identifies \emph{``pre-determined areas of concern"}. Other tools, such as Global Affairs Canada (GAC)'s AI Process Assistance Tool, overlap with informational systems but differ in that their outputs propose what should be done next rather than merely describing information. In these contexts, algorithmic recommendations modulate frontline discretion: although authority formally remains with human decision-makers, such systems can narrow interpretive latitude by nudging workers toward preferred pathways, particularly when aligned with performance metrics. 

We also identified systems that align with ADMAPS's definition of \textbf{predictive systems} that estimate risks, forecast future outcomes, or estimate likelihoods. For instance, the CBSA's  Courier Low Value Shipments uses ``probabilistic models for the risk scoring of packages", while ESDC's Employment Insurance ML Workload works on`{`identifying cases where a recalculation will not result in any change."}. Similar dynamics appear in metric-driven systems at Environment and Climate Change Canada and Fisheries and Oceans Canada, such as automated oil spill detection and oceanographic anomaly detection. By formalizing uncertainty into decision pipelines, predictive systems reshape where discretion accumulates and may encourage officers to prioritize model-flagged cases over situational judgment, a form of algorithmic over-reliance identified in ADMAPS.


Finally, several systems \textbf{automate operational tasks} to optimize workflows and delegate procedural work. For example, ECCC's OCR/LLM CITES Permit Project converts scanned documents into structured data, while agencies such as Natural Sciences and Engineering Research Canada use AI to automate data entry, detect and duplicate documents, and support grant analysis. By mechanizing tasks once handled by experienced staff, operational automation reshapes the boundary between routine and discretionary work, thus boosting efficiency but potentially eroding opportunities for tacit knowledge development that ADMAPS identifies as central to professional expertise. When automation reorganizes workload distribution, it also restructures evaluative horizons: workers may be judged on throughput rather than judgment, redefining professional identity and expectations.

\subsubsection{Degrees of Uncertainty}
Examining the Canadian AI Register entries through the lens of ADMAPS reveals a spectrum of uncertainty expressions--ranging from explicit acknowledgment, to strategic minimization, to complete omission. These patterned articulations of uncertainty shed light on how agencies anticipate human discretion, envision bureaucratic processes, and justify the deployment of algorithmic tools.

A large subset of systems \textbf{explicitly acknowledge uncertainty} through linguistic framing like \emph{``pilot," ``experiment," ``testing," ``exploring," ``intended to establish infrastructure," or ``in development"}. This positioning treats algorithmic models as provisional and exploratory, reflecting ADMAPS's insight that uncertainty should be made visible so that their primary users can calibrate their interventions. For example, systems such as Westlaw Edge with CoCounsel, Lexis+ AI, or Nuix Neo are openly framed as ``pilot only," with evaluations underway comparing AI-generated outputs to conventional human workflows. Other systems, such as Risk Scoring for CLVS, APIDS, or AI-assisted weather prediction, reflect modeling uncertainty, requiring generalization from noisy or incomplete data. Moreover, the Administrative Tribunals Support Service of Canada describes an AI system as \emph{``the required infrastructure, tool, and skill foundation"} for future AI and definitive systems to emerge. Their descriptions often include qualifiers such as ``may allow," ``exploring," or ``intended to streamline," foregrounding that, in ADMAPS terms, their ``degrees of uncertainty" in outcomes remain large and algorithmic performance cannot yet be relied upon as authoritative.



In contrast, other entries \textbf{present AI systems as stable} by framing those as mature, standardized, or evidence-driven, emphasizing improvements in ``accuracy," ``reliability," ``efficiency," or ``consistency." Systems such as Document Detective, Electronic SIN Automation, Record of Employment Comments Assessment, or Export Compliance Dashboard use language that implies performance stability, even when the underlying tasks (e.g., identity verification, anomaly detection, document comparison) are prone to classification error and contextual ambiguity. For instance, Electronic Social Insurance Number (SIN) Automation highlights its ability to verify identity documents using computer vision, suggesting near-real-time reliability. Using a blanket phrase like \emph{``when certain criteria are met,"} the entry tries to downplay the inherent uncertainty in document authentication, due to variations in image quality, emerging fraud vectors, and contextual cues. By foregrounding efficiency over limitations, these systems shift uncertainty away from the algorithm and onto human officers who must resolve edge cases after confident classifications are produced. ADMAPS warns that such a performative impression of stability risks narrowing the perceived space for human discretion. Thus, uncertainty is not eliminated, but is obscured.


A third category consists of systems whose descriptions contain \textbf{no explicit reference to uncertainty}. They reflect on no risks, no limitations, no error surfaces. This silence itself is meaningful. Many systems describe tasks involving classification, summarization, or clustering--activities that inherently entail sociotechnical uncertainty. Still, their descriptions focus solely on productivity, speed, or convenience, offering no guidance on how human discretion should be configured in light of algorithmic fallibility. This silent omission effectively treats uncertainty as external to the system and relocates responsibility to human actors without naming it as such.

\subsection{Bureaucratic Processes}
We examined how the bureaucratic processes surrounding AI systems are shaped by GC's internal and external resources and constraints, administrative scaffolds and training expectations, as well as federal and provincial laws and policies.

\subsubsection{Resources and Constraints}
We examined how the intended results of the AI systems are framed in the register. Many systems explicitly emphasize goals such as \emph{``reducing workload," ``managing backlogs," ``improving efficiency," or enabling cases to be ``attended to quickly"}. These recurring justifications position algorithmic systems as compensatory infrastructural responses to staffing shortages, procedural burdens, and chronic workload pressure. ADMAPS conceptualizes such constraints and related resources as material conditions (e.g., staffing, expertise, funding, and organizational bandwidth) that shape how algorithmic systems are adopted and integrated into bureaucratic practice.

\begin{figure}
    \centering
    \begin{tabular}{cc}
        \multicolumn{2}{c}{\includegraphics[width=\linewidth]{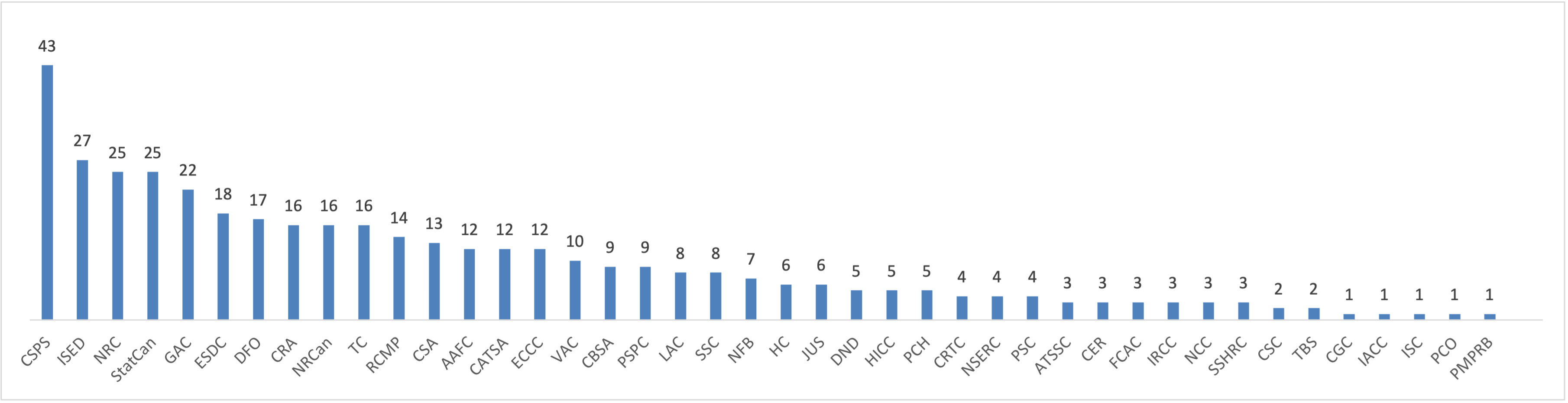}} \\
        \includegraphics[width=0.3\linewidth]{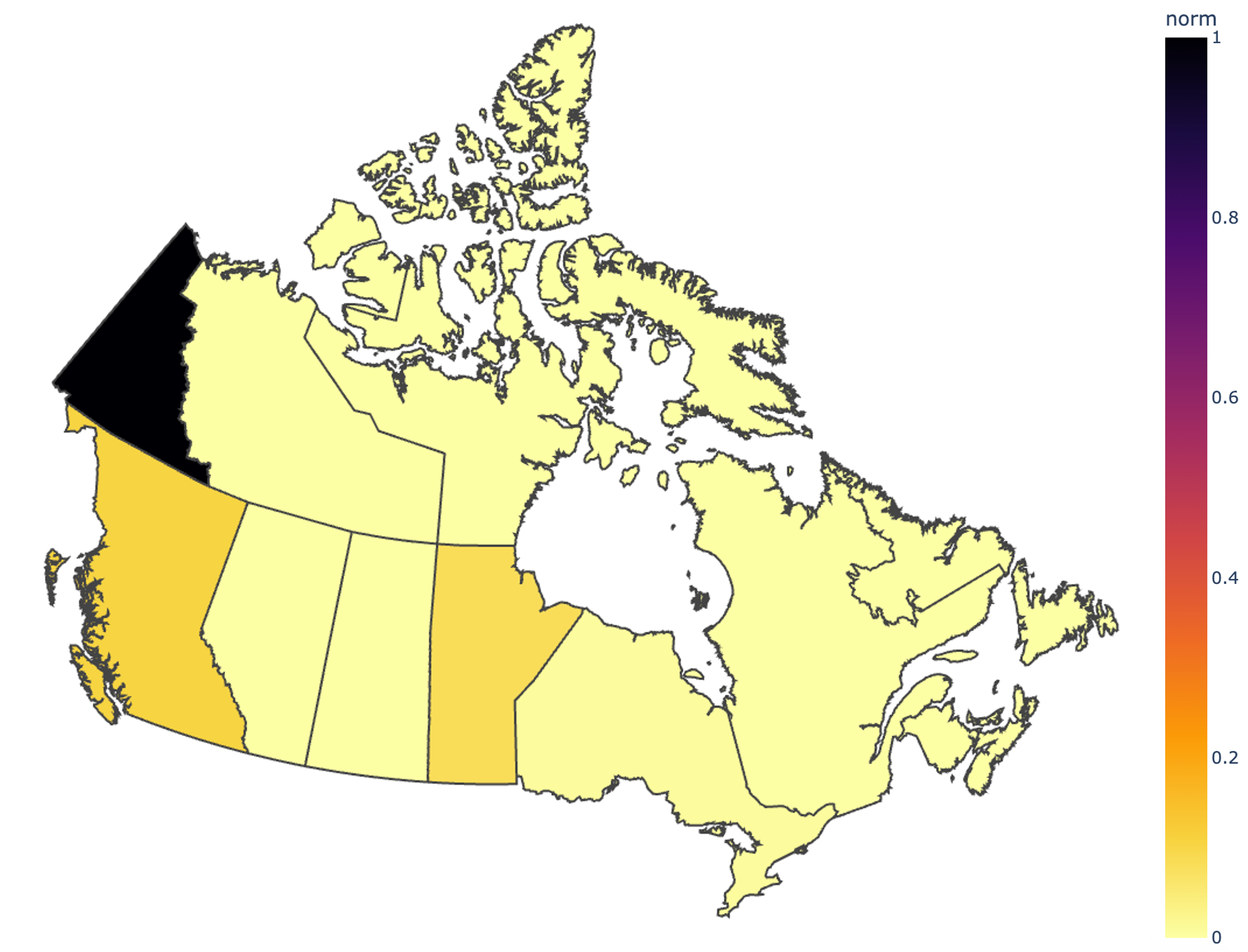} & \includegraphics[width=0.6\linewidth]{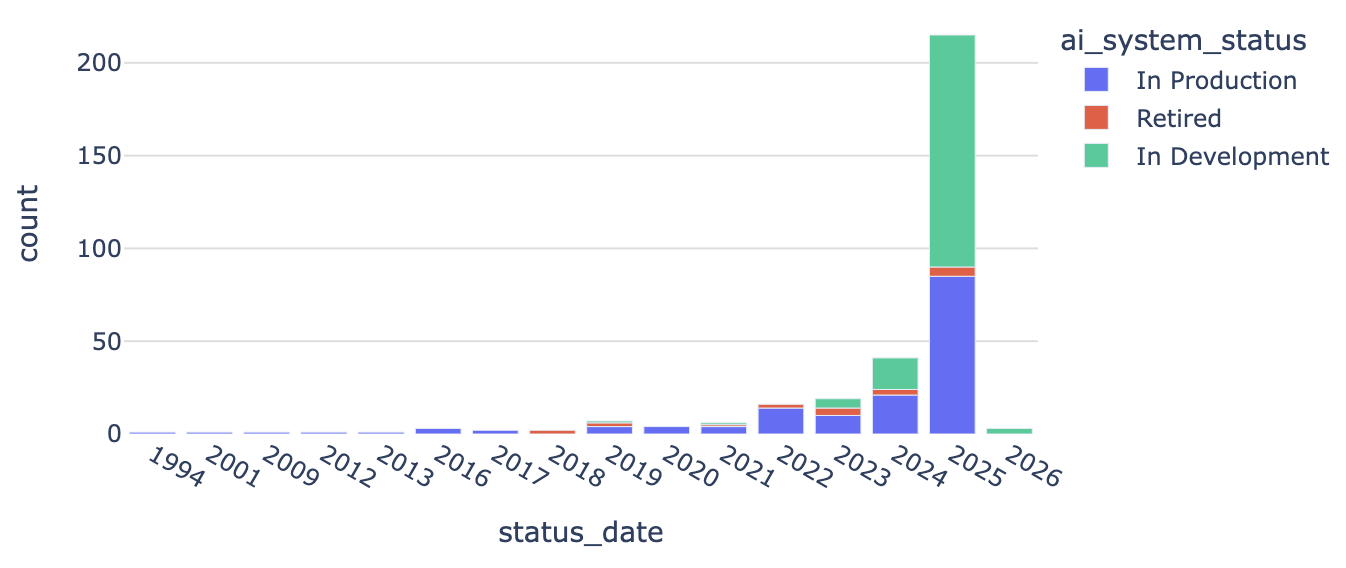}
    \end{tabular}
    \caption{Top: Distribution of AI systems across GC organizations; Bottom-left: Provinces and Territories explicitly mentioned in the AI register using population-normalized values; Bottom-right: Distribution of AI systems across years and status.}
    \label{fig:bureaucratic}
\end{figure}

There are over 200 departments and agencies in the Government of Canada~\cite{gc2025departments}, yet only 42 organizations appear on the AI register. Of which, an even smaller number of organizations, such as the Canada School of Public Service (CSPS), Innovation, Science and Economic Development Canada (ISED), National Research Council Canada (NRC), Statistics Canada, GAC, and ESDC, account for the majority of systems (see \editing{Figure~\ref{fig:bureaucratic}: top)}. Since 2022, the register shows a temporal trend of increased AI development and deployment (see \editing{Figure~\ref{fig:bureaucratic}: bottom-right)}: 44\% of systems are ``In Development," while 39\% are already ``In Production," implying simultaneous expansion and maintenance rather than isolated projects. We also observed institutional borrowing and infrastructural reuse, where systems developed within one organization inform deployments in other departments and agencies. For example, Justice Canada (JUS)'s adoption of Westlaw Edge mirrored evaluations of Lexis+ AI and Nuix Neo, while GAC's partnership with NRC enabled early deployment of scalable NLP-based solutions. ADMAPS notes that continuous system building under such conditions can limit opportunities for reflection, retraining, or recalibration of bureaucratic processes.

\begin{figure}[!ht]
    \centering
    \includegraphics[width=\linewidth]{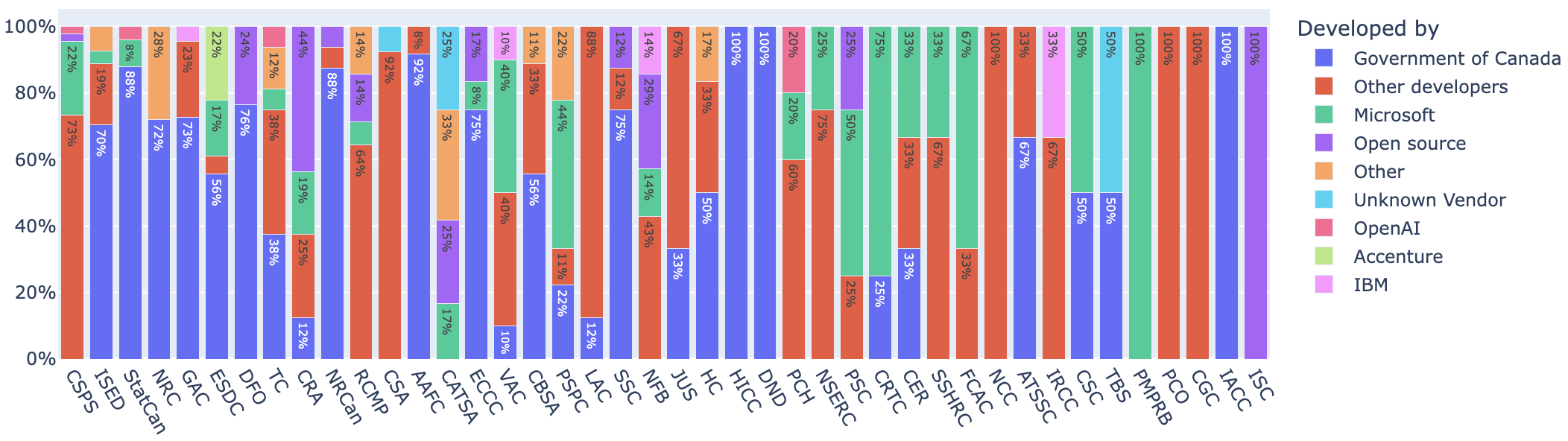}
    \caption{Proportion of developers by GC organizations.}
    \label{fig:org_dev_proportion}
\end{figure}

Patterns in system development strategies signal agencies' resource availability and their navigation of capacity constraints (see Figure~\ref{fig:org_dev_proportion}). Systems are almost evenly split between in-house development (43.3\%) and external vendors (38.1\%), while open-source development remains limited (6.1\%). A small number of vendors, most notably Microsoft (n=46), account for a disproportionate share of externally developed systems. \editing{This concentrated reliance on a small number of foreign technology providers in public-sector procurement dynamics can shape both the technological choices and governance practices surrounding public-sector AI systems.} Moreover, we identified three development strategies. First, exclusive in-house development (e.g., National Defence, Housing, Infrastructure and Communities Canada), indicating sustained internal technical expertise. Second, full reliance on external vendors or the adoption of open-source tools. And third, a mixed approach with some adopting a fully third-party approach, sometimes depending entirely on a single external vendor (e.g., Darktrace, Microsoft) or primarily on open-source systems. A third group follows a mixed strategy, combining in-house development with external vendors. For example, CSPS draws on a particularly large and heterogeneous vendor ecosystem involving 33 different entities. Under the ADMAPS framework, these differences are not merely technical choices but reflect how agencies allocate scarce resources, navigate procurement constraints, and balance short-term operational demands against longer-term capacity development.

\subsubsection{Administration and Training}
Administration and training foreground how organizational protocols, workflows, and capacity-building practices condition the operation of algorithmic systems in the public sector. Rather than treating AI systems as standalone technical artifacts, this dimension draws attention to the bureaucratic infrastructures (e.g., procedures, routinized practices, and institutional learning) that enable or constrain their use in practice.

Across the register, many systems are framed as \textbf{internal workflow instruments} rather than public-facing applications. For instance, these include service management chatbots (e.g., Milo 2.0), AI-assisted legal research platforms (e.g., Westlaw Edge), and internal generative AI tools for drafting, coding, summarization, or analysis (e.g., Genni). In these cases, AI is positioned as a means of optimizing bureaucratic throughput by \emph{reducing wait times, accelerating internal workflows, automating repetitive tasks, improving quality, or reallocating staff attention toward ``more complex" tasks}. Despite frequent references to the need for \textbf{adjusted procedures and workflows}, entries that mention systems integrating with ticketing tools, legacy databases, document repositories, or cross-departmental infrastructure often leave the new intake, handoff, and review protocols they implicitly require unexplained. 

Similarly, \textbf{training requirements} are rarely articulated. Many AI systems entail layered, multi-step interactional workflows: threshold configuration, probabilistic interpretation, large-scale document retrieval, validation of AI-generated outputs, and coordination of human-in-the-loop review~\cite{amershi2019guidelines}. AI Systems (e.g., for medical adjudication or security screening) require both expertise and familiarity with AI interfaces and limitations, yet the register rarely specifies how workers are trained to interpret outputs, recognize failures, or integrate AI with professional judgment. Training is treated as implicit rather than a formal element of algorithmic governance. Even in pilot deployments, the focus remains on anticipated outcomes rather than on institutional preparation. 

The GC AI register consistently presents AI as functional solutions, privileging a tool-centric representation of public-sector AI view that treats administrative processes and training infrastructures as peripheral or self-evident. This framing obscures the role of bureaucratic capacity and presents AI adoption as a technical intervention rather than an ongoing organizational transformation.

\subsubsection{Laws and Policies}
Laws and policies define formal constraints under which public-sector algorithmic systems are conceived, justified, and operationalized, encompassing not only legislation and regulatory mandates but also jurisdictional boundaries that allocate authority. Examining the register through the ADMAPS framework reveals that references to provinces and territories are sparse and uneven: only a subset of jurisdictions are explicitly named (see Figure~\ref{fig:bureaucratic}). The provincial or territorial references tend to appear in systems that must explicitly navigate multi-jurisdictional legal frameworks, such as regulatory compliance, enforcement, or coordination across federal–provincial–territorial boundaries. For example, tools (e.g., BizPaL) that aim to help their users navigate permits and licences across federal, provincial, and territorial regimes make jurisdiction explicit precisely because regulatory authority is fragmented. Here, AI is positioned as a mediator between overlapping legal systems, translating heterogeneous legislative requirements into standardized guidance. Conversely, the majority of AI systems in the register, particularly internal productivity tools, enterprise chatbots, and generative AI platforms, are described in jurisdiction-agnostic terms. Many of these systems are certified to process Protected B information (a classification under the GC's security framework for handling medium-sensitivity government information~\cite{microsoft2025canada}).

These systems operate within federal administrative authority, where legal constraints (e.g., privacy classification, information security, official languages requirements) are assumed to be uniform across the federal bureaucracy. \editing{This assumption becomes more complex for systems (e.g., Fuzzy Search) that rely on personal data banks while being developed or maintained by external vendors, many of whom are foreign entities. While Canada's Personal Information Protection and Electronic Documents Act~\cite{gc2000pipeda} governs private-sector processing of personal information in commercial contexts, the register provides little information on how privacy and data protection are federally governed in public-sector AI systems that rely on vendor-provided infrastructure.}

\editing{In Canada, federal responsibilities include security and national defense, whereas provinces govern domains such as health, education, and welfare, with areas like immigration and agriculture falling under concurrent jurisdiction~\cite{simeon2006division}. Though the register lists AI systems operating across several of these areas, it offers little visibility into how systems coordinate across jurisdictional domains. Instead, federal authority appears naturalized as the default governance layer, with provincial variation surfacing only when it introduces explicit regulatory complexity.} The selective mentions of provinces and territories also underscore how law shapes what must be documented in the register. While most AI systems are implicitly framed as nationally applicable, the provincial specificity of some thus signals legal complexity rather than localized deployment. This pattern reinforces a tool-centric view of legality, in which legal complexity is acknowledged only when it cannot be abstracted away. The register does not, for instance, describe how federal AI systems interact with provincial privacy regimes, First Nations governance structures, or regional interpretations of shared mandates, despite these being consequential legal realities in practice.

%

\subsection{Human Discretion}
About 86.3\% of AI systems in the register are explicitly designed for GC employees, with only a small share oriented toward public or mixed-audience applications, highlighting the centrality of human value, heuristics, and expertise.

\subsubsection{Professional Expertise}
Following ADMAPS, we examine professional expertise as tacit, experiential, and domain-specific knowledge used by public-sector practitioners to interpret policy, evidence, and cases. Across the AI register, decision-making authority remains with trained practitioners, who are expected to interpret and act on system outputs. For example, legal research and drafting tools (e.g., Westlaw Edge AI, Lexis+ AI, and Otto) are framed as assistants for JUS lawyers, while Security Screening Automation supports officers by flagging areas of concern, and final decisions explicitly remain human-led. Our analysis examines how such systems augment professional judgment and whether expertise is standardized or formalized through the system itself. Many systems are framed as mechanisms to \textbf{augment professional expertise} by reducing cognitive and administrative burdens. Generative AI tools for summarization, search, translation, and coding, such as ProteBee, Document Detective, and Science-GPT, allow practitioners to reallocate time away from clerical labor toward interpretive or discretionary work. Notably, several systems explicitly emphasize capacity-building rather than automation: ProteBee aims at\emph{``reducing the analysis processing time by at least 50\%"}, while Science-GPT is co-developed with subject matter experts to limit hallucination and ground outputs in domain knowledge, positioning AI as a scaffold for professional judgment rather than a substitute.

At the same time, many systems \textbf{encode professional expertise into structured workflows}, predefined criteria, or assessment frameworks. Risk-scoring tools, triage systems, and classification models, such as Fuzzy Search and Travelers Compliance Indicators, translate historically tacit judgments into formalized rules, features, and thresholds. While such systems promise consistency, they also shift how expertise is expressed: from situated, case-by-case reasoning to standardized representations of what matters. This encoding is particularly visible in evaluation and policy contexts. Tools like GAC's Enhanced Reporting for International Assistance explicitly embed the domain knowledge of experienced evaluators into machine-readable policy markers, producing outputs comparable to expert assessments while enabling oversight through explanations, making those structured, repeatable, and scalable across cases. \editing{However, one notable pattern across the registry entries is the limited presence of completed Algorithmic Impact Assessments, suggesting a gap between Canada's formal assessment requirements under the Directive on Automated Decision-Making~\cite{gc2021adm} and the systems publicly documented in the AI register.}

\subsubsection{Value Judgments}
We found that the key values of efficiency/speed, accuracy/quality, equity/fairness/accessibility, and risk/safety in high-stakes contexts are all deeply embedded in both the design and application of AI systems in the register. For example, the Privy Council Office's DataMinr First Alert is designed to enhance \textbf{efficiency} by summarizing large volumes of data in real time, thereby speeding up decision-making. Efficiency is often prioritized in time-sensitive contexts. However, prioritizing speed can undermine \textbf{accuracy}, a tension that systems like DeepL Pro and GitHub Copilot address by emphasizing error reduction in translation and code development--though inconsistent data practices raise questions about the reliability of these gains. For example, First Alert's ReGenAI feature relies on \emph{``publicly available data sources accessed through the internet"}. Given that online data is often tainted by communal prejudices and stereotypes~\cite{olteanu2019social}, concerns about bias and fairness arise. In particular, public-sector AI systems are expected to uphold principles of \textbf{equity, fairness, and accessibility}. For instance, Ombot was developed to enhance access to Ombuds office services while ensuring privacy and reducing barriers for employees seeking assistance. Similarly, ESDC's EquiVision aims to improve adherence to Diversity, Equity, and Inclusion (DEI) commitments by automating the analysis of DEI data and making it easier for decision-makers to address employment equity issues. Finally, values regarding \textbf{risk and safety} are prominent in AI applications for regulatory and compliance purposes. Systems such as ISED's Entity and User Behavior Analytics use machine learning to monitor activity patterns and detect anomalies, helping identify potential cybersecurity threats or insider risks. Here, the judgment centers on balancing the safety and security with the potential risk and ethical concerns of over-surveillance and privacy violations.

\subsubsection{Heuristic Decision-Making}
Heuristic decision-making refers to the rules of thumb, shortcuts, and experiential judgments that practitioners rely on when operating under uncertainty, time pressure, and resource constraints. Considering that, in public-sector contexts, such heuristics are not signs of irrationality but pragmatic adaptations to complex environments where exhaustive analysis is infeasible~\cite{march1978bounded}, we focus on how algorithmic tools become part of mechanisms such as triage, prioritization, and flagging. First, a large subset of systems explicitly frames their role as \textbf{helping staff prioritize} work. Such triaging mirrors common human heuristics, such as starting with the highest-risk cases, while formalizing them into algorithmic criteria. Analysts still decide whether to escalate a case, initiate an investigation, or dismiss a flag, but their discretion is influenced by how the space of possible actions is shaped in advance with algorithmically generated signals. Second, some systems frame their role as \textbf{supporting complex decisions} rather than filtering cases. Here, algorithmic heuristics include looking for precedents, suggesting relevant sections, and comparing with known patterns. These systems can be understood as externalizing experienced practitioner heuristics, i.e., making tacit strategies explicit, repeatable, and faster to apply. While this can reduce cognitive load and improve consistency, it also subtly reshapes how practitioners reason through cases. Third, many AI systems on the register describe their role as \textbf{flagging} instances based on anomalies or patterns. However, as discussed before, these systems do not decide outcomes; instead, they decide what deserves attention. \editing{For example, CBSA's Fuzzy Search/SSAName3, which we discussed earlier in~\ref{sec:relevant_data}, generates ranked matches and security flags. This is a high-risk system with a high impact on people, yet the register is nearly silent about how officers review or interpret these signals in practice. This omission leaves unclear whether such flags function as advisory cues, mandatory screening triggers, or informal heuristics guiding discretionary border enforcement decisions. Such} redistribution of attention is a powerful form of heuristic intervention, as it influences which cases receive scrutiny and which pass unnoticed. Overall, these interventions relocate heuristics from the point of decision or discretion to upstream moments of attention-setting.

%% file: sections/discussion.tex
\section{Discussion}
This paper speaks to current and proposed AI/algorithm registers (e.g., in European cities and transnational contexts~\cite{kaushal2024automated, murad2021beyond, haataja2020public}). Our results suggest that the Canadian AI Register enacts a technical model of algorithmic accountability, emphasizing systems' capabilities, efficiency gains, and functional outcomes while leaving administrative processes, training arrangements, and institutional capacity largely unclear. This framing presents the GC's adoption of AI as a technical choice or solution rather than an organizational transformation, which obscures the bureaucratic processes necessary to use these technologies. In terms of uncertainty, some descriptions convey experimentation, others convey stability and provisionality, while many omit discussion of it entirely. This forces frontline practitioners to infer failure modes and manage ambiguity, thus constraining discretion and increasing the system's practical burden as these tools shape upstream triage and attention. In the following section, we discuss how this study's findings align with Canada's approach to AI governance, with research within the broader FAccT community, and with public trust in governance.

\subsection{Implication for Canadian AI Governance} 
Canada organizes AI governance through a bifurcated oversight model that differentiates between public-sector and private-sector AI operations. This has resulted in a regime characterized by substantial federal investments in private-sector AI development, alongside weak and decentralized regulatory oversight frameworks~\cite{attard-frost_governance_2024, brandusescu_missed_2025, tessono_infrastructure2025}. There are no centralized laws, regulations, or frameworks; instead, a patchwork of codes, guidelines, and directives is unevenly applied to public and private sector institutions. Flagship initiatives such as the Pan-Canadian AI Strategy and the Canadian Sovereign AI Compute Strategy have prioritized research capacity, commercialization, standards development, talent attraction, significant investments in compute infrastructure, and institutional support~\cite{government_of_canada_canadian_nodate, government_of_canada_pan-canadian_2024}, positioning Canada as a global AI hub. However, their outcomes raised concerns of disproportionately benefiting industry actors, offering limited mechanisms for public accountability, and externalizing public benefits to foreign actors~\cite{brandusescu_artificial_2021, roberge_narvals_2022}.

Although public-sector AI systems are formally governed by Treasury Board Secretariat policy instruments, such as the Directive on Automated Decision-Making and the Algorithmic Impact Assessment (AIA)~\cite{scassa_administrative_2021, attard2022comments, brandusescu2022comments}, the AI register reveals substantial variation in how departments interpret, operationalize, and document these requirements. Many entries emphasize efficiency, workload reduction, and technical capability, while providing limited detail on uncertainty, human oversight, or evaluative practices. Despite requirements to complete and publish AIA results~\cite{secretariat_directive_2024}, a study found that only 4\% of 303 automated tools had publicly available government assessments~\cite{starling2025tracking}. Our study exposes the persistent gap between formal obligations and visibility in practice: even when systems are disclosed, they often omit governance-critical information, including degrees of uncertainty, training arrangements, and discretion configuration, while departmental latitude to reinterpret central directives produces heterogeneous reporting standards. Viewed against stalled efforts to regulate private-sector AI through binding legislation such as the proposed AI and Data Act~\cite{laghaei_submission_2025, tessono_ai_2022, attard-frost_ai_2025}, the GC AI Register appears as a compensatory and not an exemplary governance artifact.

Simultaneously, the register exposes the infrastructural dependencies that complicate and highlight the persistent challenges associated with Canada's claims of AI compute sovereignty. A substantial proportion of public-sector systems relies on tools developed by transnational vendors, notably Microsoft, revealing how AI deployment is embedded in the political landscape, technical expertise, global flows of capital, data, labor, compute infrastructure, and natural resources~\cite{attard-frost_ethics_2025}. Reliance on externally developed tools, as revealed by the GC's own AI repository, indicates that Canada's pursuit of sovereign AI is structurally constrained beyond domestic policy, undermining claims of AI sovereignty and technological autonomy~\cite{Brandusescu2025Transjurisdictional}. These tensions are further illustrated by partnerships emerging from Canada's sovereign compute investments, where federally funded domestic firms collaborate with foreign entities to develop a ``fully sovereign AI factory" using NVIDIA graphics processing units (GPUs) and Hewlett Packard Enterprise computing infrastructure~\cite{noauthor_telus_nodate, riehl_telus_2025}. Overall, our study highlights the breadth of infrastructural, organizational, and transnational considerations the GC must account for in shaping its AI strategies and ambitions.

\subsection{Implication for Algorithmic Accountability Research}
Based on our findings, we propose that algorithmic accountability initiatives should be understood as instruments of \textbf{ontological design}. It means that sociomaterial practices not only create objects or systems but also actively shape the kinds of beings, relationships, and ways of knowing that can exist in the world~\cite{willis2006ontological}. From this view, the GC AI register does not merely describe algorithmic systems but actively shapes which systems, decisions, and responsibilities are recognized as governable. Our analysis suggests that the register enacts a particular ontology of AI systems--one that foregrounds generative AI and LLMs while rendering other forms of AI peripheral; an ontology of AI objectives centered on efficiency and workload reduction; and an ontology of algorithmic accountability that nominally assigns responsibility to human decision-makers while obscuring how algorithmic automation conditions those decisions. Such disclosure practices ontologically impact how definitional boundaries delimit accountability. Viewed in conjunction with the ADMAPS framework, the unarticulated concerns of discretion, expertise, and uncertainty do not disappear; instead, they are shifted away from algorithmic accountability and absorbed by frontline practitioners.

These ontological effects are most visible at the boundaries of accountability artifacts, such as the GC AI register. For example, the absence of IRCC's Chinook from the register, despite its documented role in processing temporary resident visas, study and work permit applications~\cite{molnar_walls_2024, vohra_social_2023}, illustrates how definitional and classificatory choices delimit what becomes accountable. Despite widely raised concerns about Chinook~\cite{luck2025immigration, brunner_artificial_2024}, it does not appear as a named system in the Register. Whether intentionally excluded or described under a different name, this non-appearance is consequential: it renders a materially impactful form of algorithmic mediation illegible to formal governance mechanisms. From an ontological design perspective, this is not merely a gap in coverage but an active configuration of accountability. By failing to account for how AI systems are known, named, and referenced by stakeholders outside government agencies, the GC AI Register reinforces a narrow ontology of algorithmic accountability. Taken together, such disclosure infrastructures do not simply fall short of comprehensive oversight; they actively shape and are shaped by the overlap of vocabularies and terminologies through which different stakeholders recognize, contest, and govern accountability.

\subsection{Implications for Public Trust and Governance}
Beyond Canadian AI policy and algorithmic accountability research, our findings raise broader concerns about public trust and democratic governance. Public-sector AI systems increasingly operate in high-stakes domains, such as immigration, social services, and security~\cite{broeders2015datafication, eubanks2018automating, lum2016predict}, where legitimacy depends not only on procedural compliance but on citizens' ability to understand, contest, and seek redress for algorithmically mediated decisions, and hence, transparency mechanisms are central to sustaining public trust~\cite{veale2018fairness}. Our analysis complicates this assumption by showing that partial, stylized, or selective disclosure can produce the appearance of accountability without substantive guarantees. For affected individuals, such representations offer little insight into how decisions are made, where discretion resides, or how harms might be challenged. Such disclosure practices risk shifting accountability onto the public without providing the visibility or institutional means needed to exercise it, thereby weakening the democratic promise of transparency~\cite{ananny2018seeing, pasquale2015black}. These practices may normalize procedural or symbolic accountability and bolster existing research's concern about the obfuscation of data-driven decision-making tools~\cite{saxena2020human}. When registers function as endpoints rather than entry points for scrutiny, they may stabilize public trust at the surface level. However, in public institutions, where algorithmic systems directly shape access to rights, services, and protections, such arrangements risk eroding trust over time. Taken together, we argue that the public value of governance artifacts, such as the GC AI register, lies not in their mere availability but in their capacity to enable recognition, contestation, and accountability.

%% file: sections/conclusion.tex
\section{Conclusion}
In this paper, we analyzed the complete Canadian federal Public AI Register through the lens of ADMAPS, examining it as a governance artifact rather than a neutral transparency mechanism. We showed how the register produces specific bureaucratic silences, especially around discretion, infrastructural dependence, and uncertainty, and we argued that AI registers function as instruments of ontological design. On this basis, we outlined design implications for future register schemes that aim to support democratic oversight rather than only creating the appearance of transparency. Instead of viewing such artifacts as endpoints of transparency, stakeholders (e.g., policymakers and civil society actors) should engage with these as sites of ongoing accountability. For algorithmic accountability and transparency researchers, this means moving beyond evaluating whether disclosure exists or how much information it provides to sociotechnically examine how governance artifacts actively configure who and what can be held accountable.

  